\documentclass[final,3p,times,twocolumn]{elsarticle}

\usepackage{graphicx}
\usepackage{multirow}
\usepackage{amsmath,amssymb,amsfonts}
\usepackage{amsthm}
\usepackage{mathrsfs}
\usepackage[title]{appendix}
\usepackage[table,xcdraw,svgnames]{xcolor}
\usepackage{textcomp}
\usepackage{manyfoot}
\usepackage{booktabs}
\usepackage{algorithmicx}
\usepackage{algpseudocode}
\usepackage[ruled,vlined]{algorithm2e}
\usepackage{listings}
\usepackage{siunitx}
\usepackage{times}
\usepackage{epsf}
\usepackage{mathtools}
\usepackage{hyperref}
\usepackage{enumitem}
\usepackage{wrapfig}
\usepackage{subcaption}
\usepackage{comment}
\usepackage{cleveref}
\usepackage{float}
\usepackage{tabu}
\usepackage{tikz}

\usetikzlibrary{shapes.geometric, arrows}

\definecolor{LightCyan}{rgb}{0.88,1,1}

\begin{document}

\begin{frontmatter}

\title{Enhancing Brain Tumor Classification Using Vision Transformers with Colormap-Based Feature Representation on BRISC2025 Dataset}

\author[aff1]{Faisal Ahmed\corref{cor1}}
\ead{ahmedf9@erau.edu}  


\cortext[cor1]{Corresponding author}

\address[aff1]{Department of Data Science and Mathematics, Embry-Riddle Aeronautical University, 3700 Willow Creek Rd, Prescott, Arizona 86301, USA}

\begin{abstract}

Accurate classification of brain tumors from magnetic resonance imaging (MRI) plays a critical role in early diagnosis and effective treatment planning. In this study, we propose a deep learning framework based on Vision Transformers (ViT) enhanced with colormap-based feature representation to improve multi-class brain tumor classification performance. The proposed approach leverages the ability of transformer architectures to capture long-range dependencies while incorporating color mapping techniques to emphasize salient structural and intensity variations within MRI scans.

Experiments are conducted on the BRISC2025 dataset, which consists of four classes, including glioma, meningioma, pituitary tumors, and non-tumorous cases. The model is trained and evaluated using standard performance metrics such as accuracy, precision, recall, F1-score, and area under the receiver operating characteristic curve (AUC). The proposed method achieves a classification accuracy of 98.90\%, outperforming several baseline convolutional neural network models, including ResNet50, ResNet101, and EfficientNetB2. Additionally, the model demonstrates superior generalization capability with an AUC score of 99.97\%, indicating excellent discriminative performance across all classes. These results highlight the effectiveness of combining Vision Transformers with colormap-based feature enhancement for robust and accurate brain tumor classification, suggesting its potential applicability in real-world clinical decision support systems.

\end{abstract}


\begin{highlights}

\item Proposed a Vision Transformer-based framework with colormap-enhanced feature representation for brain tumor classification.

\item Utilized the BRISC2025 MRI dataset for four-class classification: glioma, meningioma, pituitary tumor, and non-tumor cases.

\item Achieved superior performance with 98.90\% accuracy and 99.97\% AUC, outperforming ResNet50, ResNet101, and EfficientNetB2.

\item Incorporated colormap visualization to enhance feature discrimination and improve model interpretability.

\item Demonstrated strong generalization capability using early stopping and robust evaluation metrics.

\end{highlights}


\end{frontmatter}



\section{Introduction}\label{sec1}

Brain tumor classification from magnetic resonance imaging (MRI) is a critical task in medical image analysis, as early and accurate diagnosis significantly influences treatment planning and patient outcomes. However, manual interpretation of MRI scans is time-consuming and prone to inter-observer variability, motivating the development of automated and reliable computer-aided diagnostic systems.

In recent years, deep learning approaches, particularly convolutional neural networks (CNNs), have achieved remarkable success in medical image classification tasks~\cite{litjens2017survey}. Architectures such as ResNet and EfficientNet have demonstrated strong performance in brain tumor classification by effectively extracting hierarchical spatial features~\cite{he2016deep,tan2019efficientnet}. Despite these advancements, CNN-based models often struggle to capture long-range dependencies and global contextual information, which are crucial for understanding complex anatomical structures in MRI images.

To address these limitations, transformer-based models have been introduced into computer vision. The Vision Transformer (ViT)~\cite{dosovitskiy2021image} has shown strong capability in modeling global relationships by leveraging self-attention mechanisms. This makes it particularly suitable for medical imaging tasks where spatial dependencies across regions are important. 

In addition, enhancing feature representation through preprocessing techniques can further improve model performance. Colormap-based transformations provide an effective way to highlight subtle intensity variations and structural patterns in MRI scans, thereby improving feature discriminability and aiding the learning process.

In this work, we propose a Vision Transformer-based framework combined with colormap-enhanced feature representation for multi-class brain tumor classification using the BRISC2025 dataset. The proposed approach aims to leverage both global contextual learning and enhanced visual representation to achieve superior classification performance.

\noindent\textbf{Our Contributions}
\begin{itemize}

\item We propose a novel framework that integrates Vision Transformers with colormap-based feature enhancement for improved brain tumor classification.

\item We utilize the BRISC2025 dataset for four-class classification, including glioma, meningioma, pituitary tumors, and non-tumorous cases.

\item We demonstrate that the proposed method outperforms state-of-the-art CNN-based models such as ResNet50, ResNet101, and EfficientNetB2.

\item We achieve high classification performance with 98.90\% accuracy and 99.97\% AUC, indicating strong discriminative capability.

\item We provide comprehensive evaluation using multiple metrics, including precision, recall, F1-score, confusion matrix, and ROC curves.

\end{itemize}

\begin{figure*}[t!]
	\centering
	
	\subfloat[\scriptsize Non-tumor MRI image.\label{fig:NT}]{
		\includegraphics[height=3.2cm,keepaspectratio]{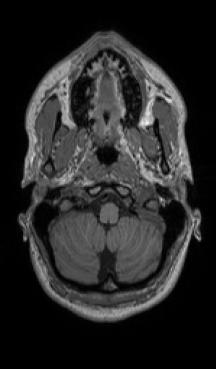}}
	\hfill
	\subfloat[\scriptsize Pituitary tumor MRI image.\label{fig:PT}]{
		\includegraphics[height=3.2cm,keepaspectratio]{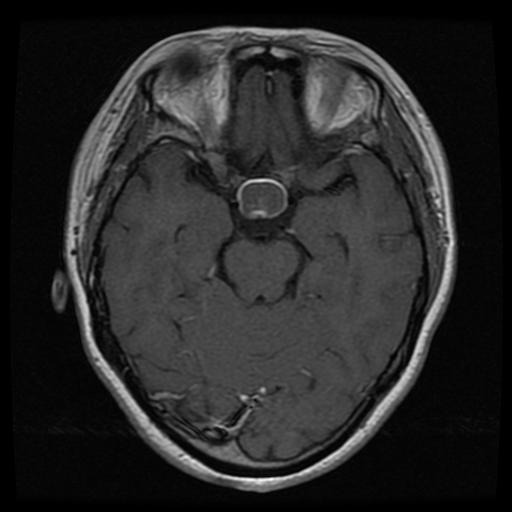}}
	\hfill
	\subfloat[\scriptsize Meningioma tumor MRI image.\label{fig:MG}]{
		\includegraphics[height=3.2cm,keepaspectratio]{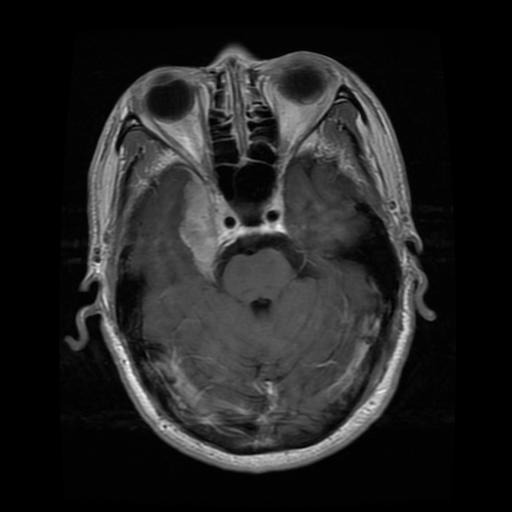}}
	\hfill
	\subfloat[\scriptsize Glioma tumor MRI image.\label{fig:GL}]{
		\includegraphics[height=3.2cm,keepaspectratio]{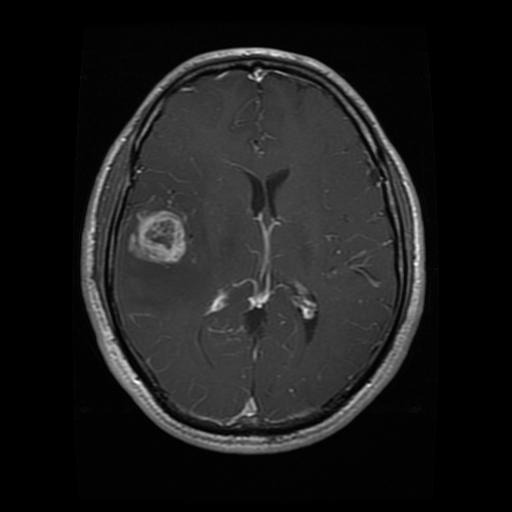}}

	\caption{\footnotesize Representative MRI samples from the BRISC2025 dataset showing the four classification categories: non-tumor, pituitary tumor, meningioma tumor, and glioma tumor. All images are resized to a uniform spatial resolution for consistent model input.}
	
	\label{fig:image-samples}
\end{figure*}

\section{Related Works}\label{sec2}

Automated analysis of brain magnetic resonance imaging (MRI) has become a key research area in computer-aided diagnosis of neurological disorders. Early studies primarily relied on handcrafted features derived from voxel-based morphometry (VBM), cortical thickness, and regional brain volumes. These features were typically combined with classical machine learning classifiers such as Support Vector Machines (SVMs), which demonstrated promising performance in distinguishing diseased subjects from healthy controls by capturing structural brain alterations~\cite{Kloppel2008SVM}. Later approaches integrated VBM features with convolutional neural networks (CNNs) to improve discriminative capability; however, these methods remained sensitive to preprocessing strategies and dataset variability~\cite{Zhang2022VBMCNN}.

The emergence of deep learning significantly advanced MRI-based classification tasks. CNN-based architectures, including 2D and 3D variants, have been widely adopted due to their ability to learn hierarchical feature representations directly from imaging data~\cite{ebrahimi2021convolutional}. More sophisticated designs, such as dense and multi-scale CNNs, further improved performance by capturing both local and contextual information~\cite{wang2021densecnn}. Ensemble learning strategies have also been explored to enhance robustness and generalization across datasets~\cite{fathi2024deep}. Despite these advances, CNN-based approaches often require large annotated datasets and extensive computational resources, limiting their effectiveness in real-world clinical scenarios with limited data availability.

To overcome these limitations, hybrid frameworks incorporating structural priors and feature selection techniques have been introduced. Topology-preserving segmentation methods, such as TOADS, enforce anatomical consistency in brain tissue segmentation~\cite{bazin2007topology}. Additionally, feature selection methods like minimum Redundancy Maximum Relevance (mRMR) have been used to reduce dimensionality and improve classification stability~\cite{alshamlan2023identifying, alshamlan2024improving}. While these approaches enhance interpretability, they are typically constrained to region-specific or intensity-based representations.

More recently, Topological Data Analysis (TDA) has emerged as a powerful tool for capturing global structural characteristics of medical imaging data. By leveraging persistent homology, TDA extracts shape-aware features that are robust to noise and invariant under small perturbations~\cite{carlsson2009topology, edelsbrunner2008persistent}. Several studies have demonstrated the effectiveness of TDA in medical imaging applications, including brain MRI and histopathological analysis~\cite{ahmed2026four, ahmed2025topo, ahmed2023topo, ahmed2023topological, ahmed2023tofi, ahmed20253d, yadav2023histopathological, ahmed2025topological, ahmed2026brain, ahmed2026hybrid}. These methods offer improved interpretability and robustness, particularly in scenarios with limited training data.

In parallel, Vision Transformers (ViTs) have gained significant attention as an alternative to CNNs for image analysis tasks. By utilizing self-attention mechanisms, ViTs effectively capture long-range dependencies and global contextual information~\cite{dosovitskiy2021image, liu2021swin}. Recent studies have explored their application in medical imaging, including MRI-based diagnosis~\cite{sankari2025hierarchical, dhinagar2023efficiently}. However, ViTs often require large-scale pretraining and may not fully exploit domain-specific characteristics of grayscale medical images. To address these challenges, several works have proposed hybrid and enhanced transformer-based approaches~\cite{ahmed2025colormap, ahmed2026hog, ahmed2025ocuvit, ahmed2025robust, ahmed2025histovit, ahmed2025addressing, ahmed2025repvit, ahmed2025pseudocolorvit, rawat2025efficient, ahmed2025transfer}.

Overall, while CNNs and transformer-based models have demonstrated strong performance, challenges such as data dependency, computational complexity, and limited interpretability remain. In contrast, topology-aware approaches provide a complementary perspective by capturing global structural patterns. Motivated by these insights, recent research trends focus on combining advanced representation learning techniques with enhanced feature extraction strategies to achieve robust and accurate classification in medical imaging applications.

\begin{figure*}[t!]
    \centering
    \includegraphics[width=\linewidth]{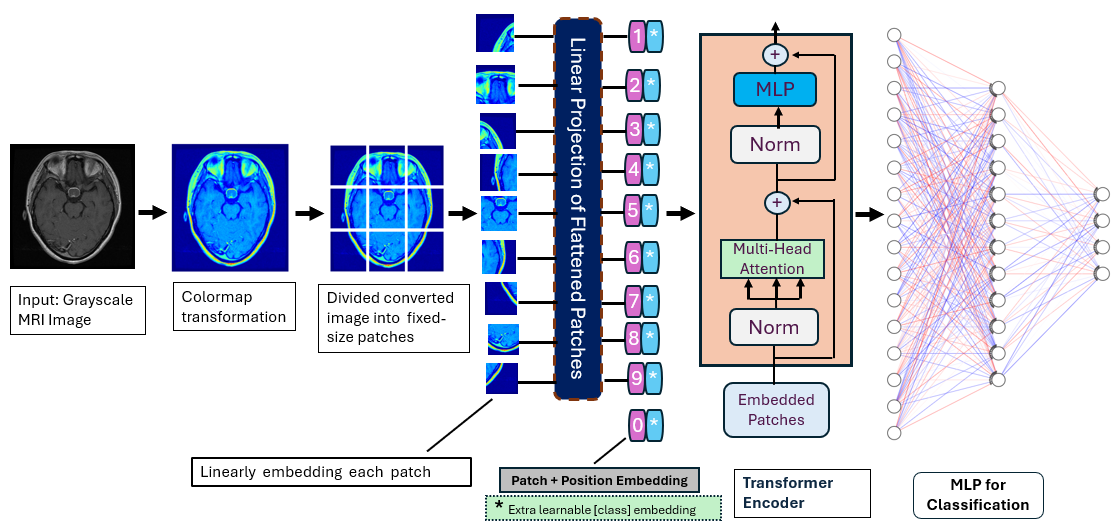}
    \caption{\footnotesize \textbf{Proposed colormap-enhanced Vision Transformer (ViT) framework for brain tumor classification.} The pipeline begins with a grayscale MRI image, which is transformed into a pseudo-color representation to enhance structural patterns and intensity variations. The color-enhanced image is resized to a fixed resolution and divided into non-overlapping patches. Each patch is flattened and linearly projected into a latent embedding space, followed by the addition of positional embeddings and a learnable classification token. The resulting sequence of embedded patches is processed through multiple Transformer encoder layers consisting of multi-head self-attention and feed-forward networks. The final representation of the classification token is passed to a multilayer perceptron (MLP) head to generate the probability distribution over the four tumor classes.}
    \label{fig:flowchart}
\end{figure*}

\section{Methodology}\label{sec:method}

This section presents the proposed methodology for four-class brain tumor classification using MRI images from the BRISC2025 dataset (\Cref{fig:image-samples}). The overall pipeline consists of image preprocessing with colormap transformation, dataset construction, Vision Transformer (ViT) fine-tuning, model optimization with early stopping, and comprehensive evaluation using classification and ROC-based metrics.

\subsection{Data Preparation and Colormap Transformation}

Let $\mathbf{I} \in \mathbb{R}^{H \times W \times 3}$ denote an input MRI image. All images are resized to a fixed resolution of $224 \times 224$ to match the input requirement of the Vision Transformer. Pixel intensities are normalized to floating-point values:
\[
\mathbf{I}_{\text{norm}} = \frac{\mathbf{I}}{255}, \quad \mathbf{I}_{\text{norm}} \in [0,1]^{224 \times 224 \times 3}.
\]

To enhance discriminative structures, a colormap transformation is applied, emphasizing intensity variations and anatomical patterns. This transformation can be expressed as:
\[
\mathbf{I}_{\text{color}} = \mathcal{C}(\mathbf{I}_{\text{norm}}),
\]
where $\mathcal{C}(\cdot)$ represents a nonlinear mapping function (e.g., jet colormap) that expands grayscale intensity distributions into a higher-dimensional color space.

The processed image is then converted into a tensor representation and permuted into channel-first format:
\[
\mathbf{X} \in \mathbb{R}^{3 \times 224 \times 224}.
\]

\subsection{Dataset Construction}

The dataset consists of four classes: glioma, meningioma, pituitary tumor, and non-tumor. Each sample is represented as a pair $(\mathbf{X}_i, y_i)$, where $y_i \in \{0,1,2,3\}$. The dataset is divided into training and testing subsets, and mini-batches of size $B=32$ are generated:
\[
\mathcal{D} = \{(\mathbf{X}_i, y_i)\}_{i=1}^{N}.
\]

Data loading is performed using a custom PyTorch \texttt{Dataset} and \texttt{DataLoader}, enabling efficient batch-wise training and shuffling.

\subsection{Vision Transformer Architecture}

We employ the pretrained Vision Transformer model \texttt{ViT-Base-Patch16-224}. Given an input image $\mathbf{X} \in \mathbb{R}^{3 \times 224 \times 224}$, it is divided into non-overlapping patches of size $16 \times 16$, resulting in $N=196$ patches:
\[
\mathbf{X} \rightarrow \{\mathbf{x}_1, \mathbf{x}_2, \dots, \mathbf{x}_N\}.
\]

Each patch is flattened and linearly projected into an embedding space:
\[
\mathbf{e}_i = \mathbf{W}_e \cdot \text{vec}(\mathbf{x}_i) + \mathbf{b}_e.
\]

A learnable classification token $\mathbf{x}_{\text{cls}}$ is prepended, and positional encodings $\mathbf{p}_i$ are added:
\[
\mathbf{z}^0 = [\mathbf{x}_{\text{cls}}, \mathbf{e}_1 + \mathbf{p}_1, \dots, \mathbf{e}_N + \mathbf{p}_N].
\]

The sequence is passed through $L$ transformer encoder layers. Each layer consists of multi-head self-attention (MSA) and a feed-forward network (FFN):
\[
\text{MSA}(\mathbf{Z}) = \text{Softmax}\left(\frac{\mathbf{QK}^\top}{\sqrt{d}}\right)\mathbf{V},
\]
\[
\mathbf{Z}' = \mathbf{Z} + \text{MSA}(\mathbf{Z}), \quad 
\mathbf{Z}^{\ell+1} = \mathbf{Z}' + \text{FFN}(\mathbf{Z}').
\]

The final representation of the classification token $\mathbf{z}_{\text{cls}}^L$ is used for prediction:
\[
\hat{\mathbf{y}} = \text{Softmax}(\mathbf{W}_c \mathbf{z}_{\text{cls}}^L + \mathbf{b}_c).
\]

\subsection{Training Strategy}

The model is trained using the categorical cross-entropy loss:
\[
\mathcal{L} = -\frac{1}{B} \sum_{i=1}^{B} \sum_{c=1}^{4} y_{i,c} \log(\hat{y}_{i,c}),
\]
where $y_{i,c}$ is the ground-truth label and $\hat{y}_{i,c}$ is the predicted probability.

Optimization is performed using the Adam optimizer:
\[
\theta_{t+1} = \theta_t - \eta \nabla_{\theta} \mathcal{L},
\]
where $\eta = 10^{-4}$ is the learning rate.

An early stopping mechanism is employed with patience $P=15$. The model parameters are updated only if validation accuracy improves:
\[
\theta^* = \arg\max_{\theta} \text{Accuracy}_{\text{val}}.
\]

\subsection{Evaluation Metrics}

Model performance is evaluated using accuracy, precision, recall, F1-score, and area under the ROC curve (AUC). Accuracy is defined as:
\[
\text{Accuracy} = \frac{1}{N} \sum_{i=1}^{N} \mathbb{I}(\hat{y}_i = y_i).
\]

Macro-averaged precision and recall are computed across all classes:
\[
\text{Precision} = \frac{1}{C} \sum_{c=1}^{C} \frac{TP_c}{TP_c + FP_c}, \quad
\text{Recall} = \frac{1}{C} \sum_{c=1}^{C} \frac{TP_c}{TP_c + FN_c}.
\]

Final predictions are obtained as:
\[
\hat{y}_i = \arg\max_c \hat{y}_{i,c}.
\]

For multi-class evaluation, the AUC is computed using a one-vs-rest (OvR) strategy:
\[
\text{AUC}_{\text{macro}} = \frac{1}{C} \sum_{c=1}^{C} \text{AUC}_c.
\]

Confusion matrices and ROC curves are generated to analyze class-wise performance and separability (\Cref{fig:auc_conf}). All experiments are implemented using PyTorch with GPU acceleration when available. The flowchart of our model is given in \Cref{fig:flowchart}.

\subsection{Hyperparameter Settings}

The performance of the proposed Vision Transformer model is influenced by several hyperparameters related to training, optimization, and architecture configuration. These parameters are carefully selected based on empirical evaluation to ensure stable convergence and optimal performance. Table~\ref{tab:hyperparameters} summarizes the key hyperparameter settings used in this study.

\begin{table}[h!]
\centering
\caption{Hyperparameter configuration for the proposed ViT-based model.}
\label{tab:hyperparameters}
\setlength{\tabcolsep}{6pt}
\begin{tabular}{ll}
\toprule
\textbf{Hyperparameter} & \textbf{Value} \\
\midrule
Model Backbone & ViT-Base-Patch16-224 \\
Input Image Size & $224 \times 224$ \\
Patch Size & $16 \times 16$ \\
Number of Classes & 4 \\
Batch Size & 32 \\
Number of Epochs & 50 \\
Optimizer & Adam \\
Learning Rate & $1 \times 10^{-4}$ \\
Loss Function & Cross-Entropy Loss \\
Early Stopping Patience & 15 \\
Weight Initialization & Pretrained (ImageNet) \\
Device & GPU/CPU (Auto-detected) \\
Shuffle (Training) & True \\
\bottomrule
\end{tabular}
\end{table}

\begin{algorithm}[t]
\SetAlgoNlRelativeSize{0}
\DontPrintSemicolon
\caption{PseudoColorViT: Colormap-Enhanced Vision Transformer for MRI Classification}
\label{alg:pcvit}

\KwIn{MRI dataset $\mathcal{D}=\{(\mathbf{I}_i,y_i)\}_{i=1}^{N}$, pretrained ViT, epochs $E$, patience $p$, classes $C=4$}
\KwOut{Best model and evaluation metrics}

\textbf{Preprocessing:}
\For{$i=1$ \KwTo $N$}{
Resize $\mathbf{I}_i \to 224\times224$;
Apply colormap $\mathbf{I}_i \gets \mathcal{C}(\mathbf{I}_i)$;
Normalize $\mathbf{I}_i \gets \mathbf{I}_i/255$;
Convert to tensor $\mathbf{X}_i \in \mathbb{R}^{3\times224\times224}$;
}

Split dataset into train/test (80:20);

\textbf{Model Setup:}
Load ViT-Base-Patch16-224;
Replace classifier with $C$-class Softmax;
Initialize Adam optimizer ($\eta=10^{-4}$);
Define cross-entropy loss;

\textbf{Training:}
Initialize best accuracy $\alpha^*=0$, counter $\delta=0$;

\For{epoch $=1$ \KwTo $E$}{
Train model on mini-batches $(\mathbf{X},y)$ using forward pass and backpropagation;

Evaluate on test set;

\If{accuracy $>\alpha^*$}{
Save model; update $\alpha^*$; reset $\delta=0$;
}
\Else{
$\delta \gets \delta+1$;
}

\If{$\delta \ge p$}{
break;
}
}

\textbf{Evaluation:}
Load best model;
Compute predictions using Softmax;
Calculate Accuracy, Precision, Recall, F1-score, AUC;
Generate confusion matrix and ROC curves;

\Return trained ViT model and performance metrics;

\end{algorithm}
\section{Experiment}

\subsection{Datasets}

\noindent \textbf{BRISC2025 Dataset}~\cite{brisc2025} \\
The BRISC2025 dataset is a publicly available benchmark designed for brain tumor segmentation and classification tasks using magnetic resonance imaging (MRI). It consists of approximately 6,000 contrast-enhanced T1-weighted MRI scans that have been carefully annotated by certified radiologists and medical experts. The dataset includes four categories: three tumor types—glioma, meningioma, and pituitary tumor—as well as non-tumorous cases. 

Each sample is accompanied by high-resolution pixel-level segmentation masks, enabling both classification and fine-grained segmentation analysis. Additionally, the dataset is organized across multiple anatomical imaging planes, including axial, sagittal, and coronal views, which supports robust model generalization and cross-view learning. 

To facilitate model development and evaluation, the dataset is typically divided into training and testing subsets, with around 5,000 images for training and 1,000 for testing. The high-quality annotations and balanced class representation make BRISC2025 a valuable resource for developing and benchmarking deep learning models in neuro-oncological image analysis.

\subsection{Experimental Setup}
\noindent \textbf{Training–Test Split:} We utilize the predefined training and testing partitions provided with the BRISC2025 dataset. This standardized split ensures fair comparison with existing studies and preserves consistency in evaluation protocols for four-class brain tumor classification.
\smallskip

\noindent \textbf{No Data Augmentation:}
Unlike many deep learning approaches that rely on extensive data augmentation techniques to address limited or imbalanced training samples~\cite{goutam2022comprehensive}, the proposed colormap-enhanced Vision Transformer framework is trained without applying any data augmentation. The use of pretrained transformer representations combined with colormap-based feature enhancement improves the model’s ability to capture discriminative structural patterns directly from the original MRI images. 

The self-attention mechanism in the Vision Transformer captures global contextual relationships, while the colormap transformation enhances subtle intensity variations that are informative for tumor classification. As a result, the model demonstrates strong generalization capability without requiring artificial transformations such as rotation, flipping, or scaling. This simplifies the training pipeline, reduces computational complexity, and preserves the original anatomical characteristics of MRI scans.

\noindent \textbf{Runtime Efficiency and Platform:} 
All experiments were conducted on a personal laptop equipped with an Apple M1 system-on-chip, featuring an 8-core CPU (4 performance cores and 4 efficiency cores) and 16~GB of unified memory. 

\section{Results}\label{sec:results}

The performance of the proposed Vision Transformer (ViT) with colormap-enhanced feature representation is evaluated on the BRISC2025 dataset for four-class brain tumor classification. The model is assessed using standard evaluation metrics, including accuracy, precision, recall, F1-score, and area under the ROC curve (AUC). A comparative analysis with state-of-the-art convolutional neural network (CNN) models is also conducted to demonstrate the effectiveness of the proposed approach.

Table~\ref{tab:results} summarizes the quantitative results. Among the baseline models, ResNet50 achieves an accuracy of 98.20\%, while ResNet101 records 98.09\%. EfficientNetB2 slightly improves performance with an accuracy of 98.37\%. Although these CNN-based architectures demonstrate strong classification capability, their performance remains marginally lower compared to the proposed method.

In contrast, the proposed ViT-based model achieves the highest accuracy of 98.90\%, along with a precision of 98.99\%, recall of 98.98\%, and F1-score of 98.98\%. This consistent improvement across all evaluation metrics indicates the robustness and reliability of the proposed framework. Notably, the model attains an AUC score of 99.97\%, demonstrating excellent discriminative ability across all four classes.

The superior performance of the proposed method can be attributed to two key factors. First, the Vision Transformer effectively captures global contextual relationships within MRI images through its self-attention mechanism, overcoming the locality limitations of CNNs. Second, the incorporation of colormap-based feature representation enhances the visibility of subtle intensity variations and structural patterns, leading to improved feature separability.

\begin{figure*}[t!]
    \centering
    \subfloat[\scriptsize One-vs-rest ROC curves with corresponding AUC scores for each class.\label{fig:auc}]{
        \includegraphics[width=0.45\linewidth]{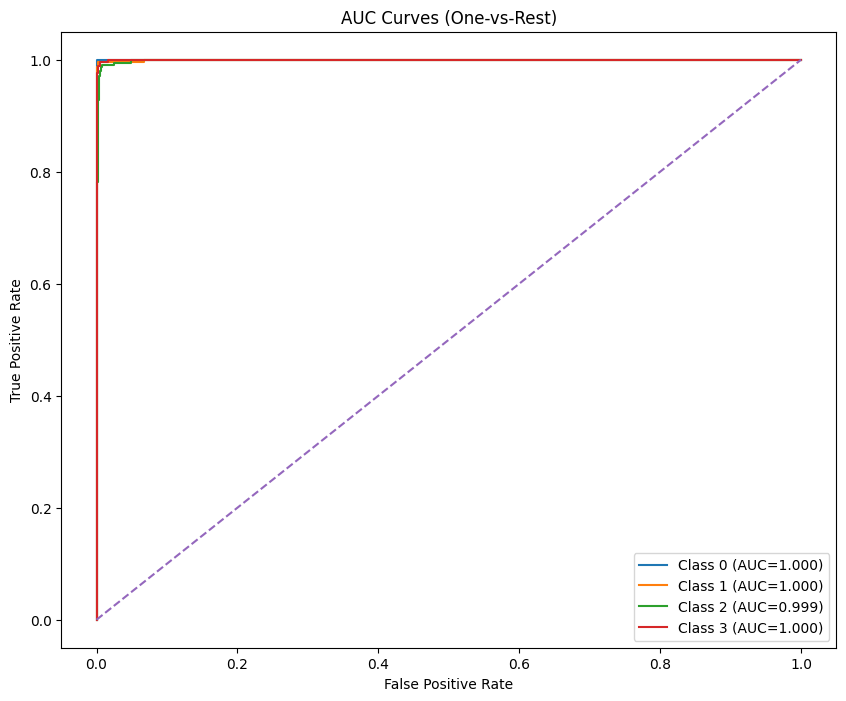}}
    \hfill
    \subfloat[\scriptsize Confusion matrix showing class-wise prediction performance for four-class Alzheimer’s disease classification.\label{fig:conf}]{
        \includegraphics[width=0.45\linewidth]{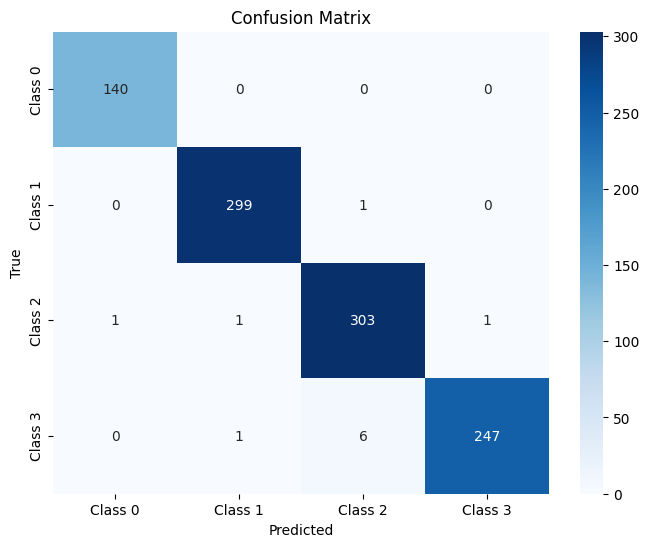}}
    
    \caption{\footnotesize Evaluation of \textbf{TDA+DenseNet121} on the OASIS-1 dataset. (a) One-vs-rest ROC curves illustrating discrimination performance across the four Alzheimer’s disease classes. (b) Confusion matrix showing correct and misclassified instances for each class.}
    \label{fig:auc_conf}
\end{figure*}

Further analysis using the confusion matrix reveals that the model achieves high classification accuracy across all classes, with minimal misclassification (\Cref{fig:conf}). The ROC curves also indicate strong separability (\Cref{fig:auc}), as evidenced by the near-perfect AUC values for each class in the one-vs-rest setting.

Overall, the experimental results demonstrate that the proposed approach not only outperforms traditional CNN-based models but also provides a more robust and generalized solution for multi-class brain tumor classification. These findings highlight the potential of combining transformer-based architectures with enhanced feature representation techniques for advanced medical image analysis.

\begin{table*}[h!]
\centering
\caption{\footnotesize 
Published accuracy results for four-class Alzheimer’s disease classification 
on OASIS or OASIS-derived MRI datasets. 
\label{tab:results}}
\setlength\tabcolsep{4pt}
\footnotesize

\begin{tabular}{lccccccc}
\multicolumn{8}{c}{\bf OASIS / OASIS-derived MRI Dataset: 4-Class Classification Results} \\
\toprule
Method & \# Classes & Dataset & Prec & Rec & Accuracy & F1-score & AUC \\
\midrule


ResNet50~\cite{fateh2026brisc} 
& 4 & BRISC2025 & 98.36& 98.36& 98.20 & 98.34 & - \\

ResNet101~\cite{fateh2026brisc} 
& 4 & BRISC2025 &98.20 &98.26 & 98.09 & 98.21 & - \\

EfficientNetB2~\cite{fateh2026brisc} 
& 4 & BRISC2025 &9.851 &98.54 & 98.37 &  98.52& - \\

\midrule
\bf Ours
& 4 & BRISC2025 & \textbf{98.99} & \textbf{98.98} & \textbf{98.90} & \textbf{98.98} & 99.97 \\
\bottomrule
\end{tabular}
\end{table*}

\section{Discussion}\label{sec:discussion}

The experimental results demonstrate that the proposed Vision Transformer (ViT) with colormap-enhanced feature representation achieves superior performance compared to conventional CNN-based architectures. The consistent improvement across accuracy, precision, recall, and F1-score indicates that the model effectively captures both global and discriminative features from MRI data.

One of the key strengths of the proposed approach lies in the ability of the Vision Transformer to model long-range dependencies through self-attention mechanisms. Unlike CNNs, which primarily focus on local receptive fields, the transformer architecture captures global contextual relationships that are crucial for identifying complex structural patterns in brain MRI images. This contributes significantly to the improved classification performance observed in the results.

The integration of colormap-based feature representation further enhances the model’s capability by highlighting subtle intensity variations and structural details that may not be easily distinguishable in grayscale images. This preprocessing step improves feature separability and supports more effective learning, particularly in challenging multi-class classification scenarios.

Additionally, the high AUC value obtained by the proposed model reflects strong class separability and robustness. The confusion matrix analysis shows minimal misclassification across categories, suggesting that the model generalizes well across different tumor types. The use of early stopping during training also helps prevent overfitting, ensuring stable and reliable performance.

Despite these promising results, certain limitations remain. The model relies on pretrained transformer architectures, which may introduce computational overhead and dependency on large-scale pretraining. Furthermore, while the BRISC2025 dataset provides high-quality annotations, evaluating the model on additional external datasets would further validate its generalizability in real-world clinical settings.

Overall, the findings highlight the effectiveness of combining transformer-based learning with enhanced feature representation techniques for medical image classification tasks.

\section{Conclusion}\label{sec:conclusion}

In this study, we presented a Vision Transformer-based framework enhanced with colormap feature representation for multi-class brain tumor classification using MRI data. The proposed approach leverages the global context modeling capability of transformers along with improved visual feature representation to achieve high classification performance. Experimental evaluation on the BRISC2025 dataset demonstrates that the proposed model outperforms several state-of-the-art CNN-based architectures, achieving an accuracy of 98.90\% and an AUC of 99.97\%. These results confirm the effectiveness of the proposed method in capturing complex structural patterns and improving classification reliability. The study highlights the potential of transformer-based models in medical imaging, particularly when combined with appropriate preprocessing strategies such as colormap enhancement. This combination enables better feature discrimination and contributes to improved model generalization.

For future work, the proposed framework can be extended by incorporating multi-modal imaging data, exploring lightweight transformer architectures for reduced computational cost, and validating performance on larger and more diverse clinical datasets. Such advancements could further enhance the applicability of the proposed method in real-world clinical decision support systems.


 \section*{Declarations}

 \textbf{Funding} \\
 The author received no financial support for the research, authorship, or publication of this work.

 \vspace{2mm}
 \textbf{Author's Contribution} \\
 Faisal Ahmed conceptualized the study, downloaded the data, prepared the code, performed the data analysis and wrote the manuscript. Faisal Ahmed reviewed and approved the final version of the manuscript. 

  \vspace{2mm}
 \textbf{Acknowledgement} \\
The authors utilized an online platform to check and correct grammatical errors and to improve sentence readability.

 \vspace{2mm}
 \textbf{Conflict of interest/Competing interests} \\
 The authors declare no conflict of interest.

 \vspace{2mm}
 \textbf{Ethics approval and consent to participate} \\
 Not applicable. This study did not involve human participants or animals, and publicly available datasets were used.

 \vspace{2mm}
 \textbf{Consent for publication} \\
 Not applicable.

 \vspace{2mm}
 \textbf{Data availability} \\
 The datasets used in this study are publicly available online. 

 \vspace{2mm}
 \textbf{Materials availability} \\
 Not applicable.



\bibliographystyle{elsarticle-num-names}

\bibliography{refs}

\end{document}